\documentclass[letterpaper, 10 pt, conference]{ieeeconf}  

\IEEEoverridecommandlockouts                              

\usepackage{glossaries}
\usepackage{graphicx}
\usepackage{svg}
\usepackage{multirow}
\usepackage{fontawesome5}
\usepackage{booktabs}
\usepackage{xcolor}
\usepackage{amssymb}
\usepackage{algorithm}
\usepackage{algorithmicx}
\usepackage{algpseudocode}
\usepackage{url}
\usepackage{mathtools}       
\newcommand{\pluseq}{\mathrel{+}=}
\usepackage[dvipsnames]{xcolor}
\usepackage{balance}

\makeglossaries

\newglossaryentry{VLA}{
  name=Vision Language Action Models (NeRFs),
  description={A document preparation system}
}

\title{\LARGE \bf
EveryDayVLA: A Vision-Language-Action Model for Affordable Robotic Manipulation
}

\author{Samarth Chopra, Alex McMoil, Ben Carnovale, Evan Sokolson, Rajkumar Kubendran, and Samuel Dickerson
\thanks{
We gratefully acknowledge the support of the University of Pittsburgh Center for Research Computing funded by the National Institutes of Health (NIH) under NIH award number S10OD028483.
}
\thanks{
Samarth Chopra, Alex McMoil, Ben Carnovale, Evan Sokolson, Rajkumar Kubendran, and Samuel Dickerson are with the University of Pittsburgh,
    {\tt\small \{sac345, alm470, bencarnovale, evs44, rajkumar.ece, dickerson\}@pitt.edu}}%
}

\begin{document}

\maketitle
\thispagestyle{empty}
\pagestyle{empty}

\begin{abstract}

While Vision–Language–Action (VLA) models map visual inputs and language instructions directly to robot actions, they often rely on costly hardware and struggle in novel or cluttered scenes. We introduce EverydayVLA, a 6-DOF manipulator that can be assembled for \$300, capable of modest payloads and workspaces. A single unified model jointly outputs discrete and continuous actions, and our adaptive-horizon ensembler monitors motion uncertainty to trigger on-the-fly replanning for safe, reliable operation. On LIBERO, EverydayVLA matches state-of-the-art success rates, and in real-world tests it outperforms prior methods by 49\% in-distribution and 34.9\% out-of-distribution. By combining a state-of-the-art VLA with cost-effective hardware, EverydayVLA democratizes access to a robotic foundation model, and paves the way for economical use in homes and research labs alike.
Experiment videos and more details can be found on our project page: \textcolor{RoyalBlue}
{\url{https://everydayvla.github.io/}}

\end{abstract}

\section{INTRODUCTION}
Vision–Language–Action (VLA) models \cite{brohan2022rt} have transformed robotics by learning a direct mapping from raw images and natural‐language instructions to motor commands, bypassing hand‐designed perception or planning modules. Building atop large Vision–Language Models (VLMs) \cite{chen2022pali} pre‐trained on massive web‐scale datasets, VLAs exhibit impressive scene understanding and instruction following. Yet, even with internet‐scale pretraining, they remain brittle under unfamiliar lighting \cite{chi2023diffusion}, novel objects \cite{xie2024decomposing}, and visual distractors \cite{team2024octo}, and often fail to generalize to out‐of‐distribution tasks \cite{brohan2023rt}.

Currently, state‐of‐the‐art robotic manipulators typically cost thousands of dollars, driven by high‐precision actuators, custom‐machined components \cite{quigley2011low}, and multiple control boards and motor drivers \cite{arduino2015arduino} \cite{PCA9685}. This high hardware cost and system complexity—coupled with the tedious, expensive process of teleoperated data collection for fine‐tuning models \cite{dass2022pato}—severely limit accessibility and slow progress toward wider adoption \cite{christensen2015innovator}.

To address these challenges, we present a full‐stack system, and present \textbf{three distinct contributions}.

\begin{figure}[t!]
    \centering
\includegraphics[width=0.48\textwidth]{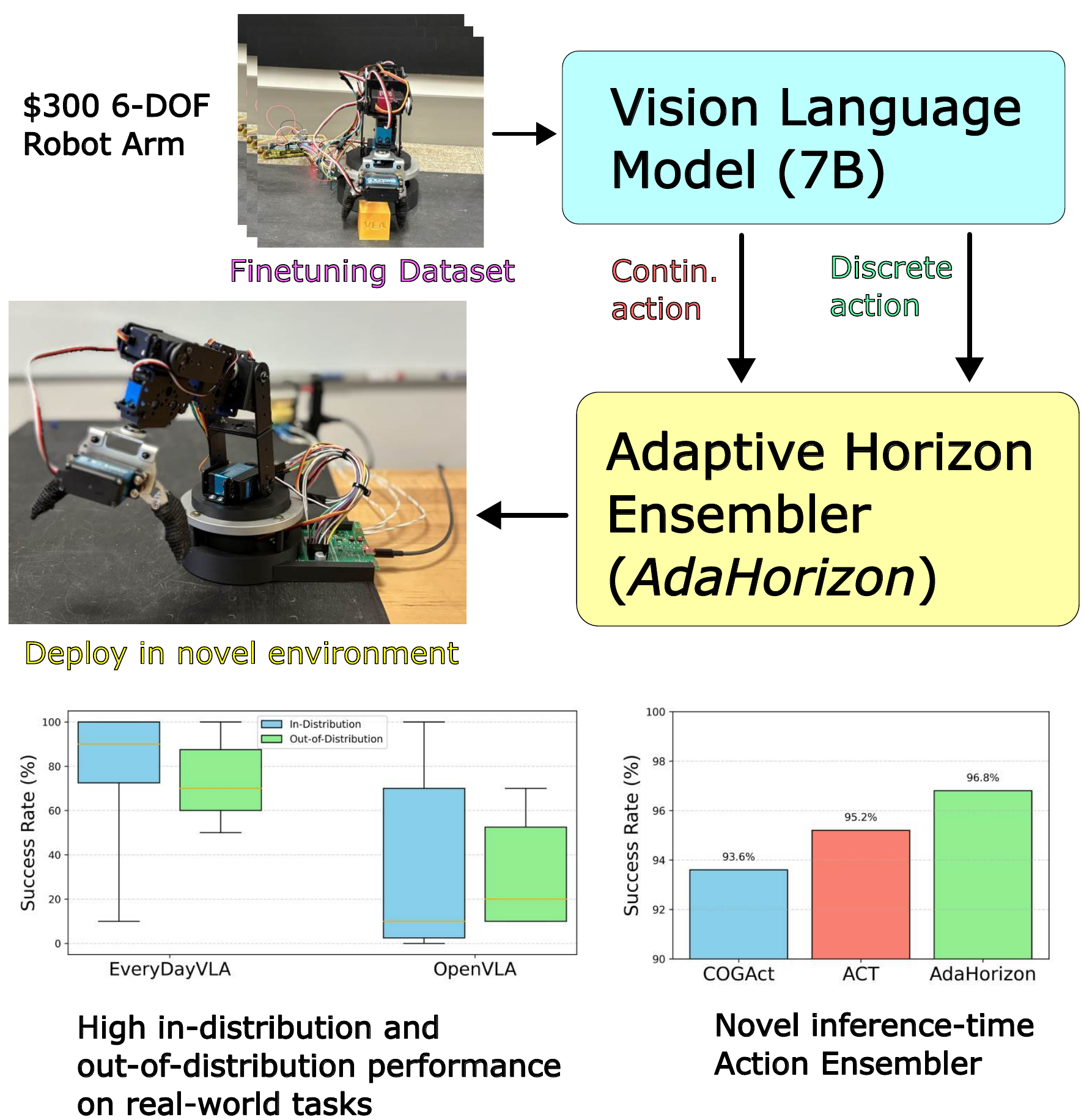}
\caption{
\textit{EveryDayVLA} system.
\underline{\textbf{Top}}: EveryDayVLA finetunes a VLA for a low-cost manipulator to generate continuous and discrete actions, which are passed to an adaptive horizon ensembler, to produce adaptive-sized action chunks, accounting for model uncertainty. 
\textbf{\underline{Bottom}}: Our model is able to show high in-distribution and out-of-distribution performance on real world tasks, and our action ensembler (\textit{AdaHorizon}) beats other state-of-art action ensemblers.
}
\vspace{-.7cm}
\label{fig:figure1}
\end{figure}

\begin{itemize}

\item Collaborative training with adaptive horizon control (\textit{AdaHorizon}). We jointly train continuous (L1-regression based) and discrete autoregressive action heads, use their disagreement to estimate model uncertainty, and dynamically adjust action horizons to trigger replanning under tight real‐time constraints.

\item A low‐cost, integrated 6-DOF manipulator. Our \$300 design achieves better than 10 mm repeatability while utilizing an accessible Arduino Uno breakout board with PCA9685 pulse with modulation (PWM) driver for 12-bit PWM control.

\item An automated data‐collection pipeline and public dataset. We streamline teleoperation to collect trajectories with language instructions, video, and end-effector poses, releasing more than 1,200 task executions to enable scalable fine‐tuning across diverse environments.

\end{itemize} 

\textit{EveryDayVLA} combined with our novel adaptive horizon algorithm (\textit{AdaHorizon}) remains competitive on the LIBERO simulation benchmark \cite{liu2023libero} and beats state-of-the-art by 34.9\% and 49\% on real‐world in-distribution and out-of-distribution scenarios respectively. By driving down hardware cost (see Table \ref{tab:lowcost-arms}), our system democratizes access to robotic foundation models, paving the way for broader research and adoption.

\begin{table*}[ht]
  \centering
  \caption{Comparison of low-cost research robotic manipulators alongside a commercial baseline.}
  \label{tab:lowcost-arms}
  \resizebox{\textwidth}{!}{
  \begin{tabular}{@{} l c c c c c c@{}}
    \toprule
    \textbf{Arm (Ref.)}                        & \textbf{Cost}   & \textbf{DOF} & \textbf{Payload} & \textbf{Workspace}      & \textbf{Max Speed} & \textbf{Repeat.}                        \\ 
    \midrule
    Low‐cost compliant manipulator  
      \cite{quigley2011low}                    & \$4,135         & 7            & 2 kg             & –                       & 1.5 m/s            &  $< 3$ mm           \\
    REPLAB platform (WidowX‐250)  
      \cite{yang2019replab}                    & \$2,000         & 6            & 0.25 kg          & 70×40×60 cm³            & –                  & ±1mm                \\
    Franka Emika Panda  
      \cite{franka}               & \$29,900        & 7            & 3 kg             & 855 mm reach            & 2 m/s              & ±0.1mm                    \\
    
    ARMADA  
      \cite{kim2025design}                     & \$3,040              & 6            & 2.5kg                & –                       & 6.16 m/s                  & 2.63mm            \\
      AhaRobot  
      \cite{cui2025aharobot}                     & \$1,000              & 14            & 2.5kg                & –                       & 6.16 m/s                  & 2.63mm
      \\
      PAMY2 \cite{guist2023safe}
                                               & \$15,835            & 4            & –                     & –                        & 12 m/s                       & –                     \\
      BLUE \cite{gealy2019quasi}
                                               &$
                                               < \$ 5,000$             & 7            & 2 kg                    & –                        & 2.1 m/s                      & 3.7mm                  \\
    
    Ours                     & \$311.98               & 6            & 0.2 kg                & 382 mm reach                       & 0.7 m/s                  & $\leq 10 $mm                                \\
    \bottomrule
  \end{tabular}
  }
\end{table*}

\section{Related Work}

\subsection{Vision Language Action (VLA) Models}


Recent advances have transformed Vision–Language Models (VLMs) into Vision–Language–Action (VLA) systems that directly predict low‐level control commands for robotic manipulators \cite{brohan2022rt}, \cite{chi2023diffusion},  \cite{brohan2023rt}, \cite{kim2024openvla}, \cite{li2023vision}, \cite{black2024pi0}, \cite{intelligence2504pi0}. 
By harnessing internet‐scale pretraining \cite{touvron2023llama} and expansive cross‐embodiment datasets \cite{o2024open}, \cite{walke2023bridgedata}, \cite{khazatsky2024droid}, these models achieve both language comprehension \cite{zhou2023language, shi2025hi} and rich scene understanding. Early VLA formulations borrowed the autoregressive “next‐token prediction” paradigm from language modeling \cite{brohan2022rt, brohan2023rt, kim2024openvla}, successfully encoding complex behaviors but struggling to learn high‐frequency, dexterous skills from dense control data \cite{pertsch2025fast}. To overcome this, recent work has shifted toward generating continuous action trajectories via diffusion and flow‐matching methods \cite{chi2023diffusion, black2024pi0, bjorck2025gr00t, zhao2024aloha, wen2025tinyvla}. Although diffusion‐based approaches can deliver higher inference throughput, they introduce notable computational trade‐offs—namely slower training \cite{zhang2024improving, wang2023patch} and multiple denoising or integration steps at inference \cite{kim2025fine, ho2020denoising}.


Tokenization‐based methods like FAST convert continuous trajectories into compact tokens, yielding up to 5× faster training than diffusion‐based VLAs \cite{pertsch2025fast}. However, because FAST still decodes tokens autoregressively, its inference latency remains high. OpenVLA-OFT \cite{kim2025fine} tackles this by combining parallel decoding, action chunking \cite{zhao2023learning}, and an L1-regression objective: the model generates entire action chunks in one forward pass, boosting throughput proportionally to chunk size. Despite these gains, OpenVLA-OFT inherits the prolonged training cycles typical of continuous‐action architectures, and in real‐world tests its FiLM‐based language grounding \cite{perez2018film} can be inconsistent.

Naturally, a hybrid approach, leveraging both autoregressive and continuous predictions have been formulated \cite{liu2025hybridvla}, with action ensembling to fuse predictions. While this has displayed strong performance and robustness on simulation and real-world tasks, it is bottlenecked by slow autoregression.

We take inspiration from this framework, and make a simple change by replacing the diffusion objective with an L1-regression objective. Further, instead of predicting one action at a time, via our autoregressive actions, we borrow OpenVLA-OFT's \cite{kim2025fine} design philosophy by leveraging action chunking and parallel decoding, enabling joint prediction of discrete and continuous action chunks. We then use the disagreement between the two predictions, to adaptive tune our action horizon at inference, embedding replanning capabilities into the model, while maintaining high inference rates.

\subsection{Low-cost Robotic Manipulators}



Several prior efforts have aimed to reduce the cost barrier to robotic manipulators for broader adoption \cite{quigley2011low, yang2019replab, cui2025aharobot, wu2024gello, cocota2012low}. However, commercially available “low-cost” arms still retail for well over \$1,000 (Table \ref{tab:lowcost-arms}), placing them out of reach for many home users and student researchers. Designing an affordable manipulator requires careful trade-offs among workspace, degrees of freedom (DOF), payload capacity, speed, and repeatability. Most sub-\$1,000 solutions compromise one or more of these critical specifications, limiting their utility in research and education. Furthermore, dependence on specialized software frameworks, such as custom driver stacks \cite{bruyninckx2001open} or ROS extensions \cite{quigley2009ros}, creates an additional barrier for novice users.











Guided by these priorities, we engineered an open-source, 6-DOF robotic arm for around \$300 that delivers a 0.2 kg payload, 382 mm reach, up to 0.7 m/s end-effector speed, and 10 mm repeatability. To maximize accessibility, we’ve minimized software dependencies and ensured full OS-agnostic support, lowering the barrier for both home enthusiasts and academic users.

\section{Method}

\begin{figure*}
    \centering
\includegraphics[width=1.0\textwidth]{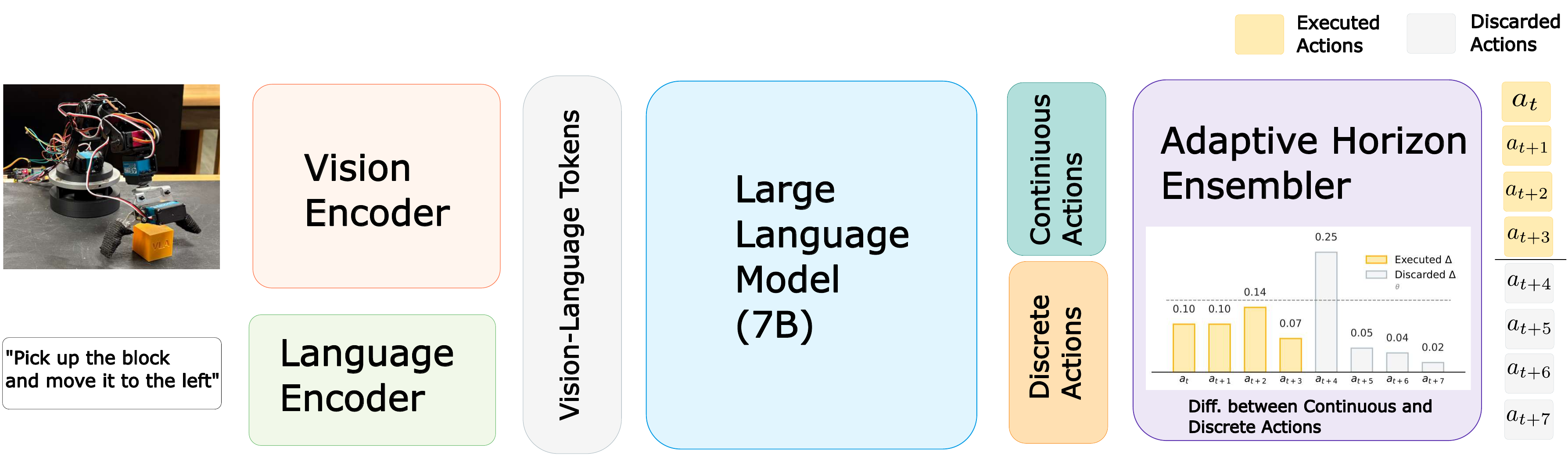}
\caption{\textit{EveryDayVLA} architecture. The VLA takes as input an image and natural language instruction and these are tokenized via the vision and language encoders, and sent to the Llama 2 LLM, which produces continuous and discrete actions. These actions are then passed to the adaptive horizon ensembler, which computes the difference between the two actions, executing those only below a certain threshold.}
\end{figure*}

Autoregressive VLAs suffer due to iterative generation, leading to danger of compounding errors \cite{bachmann2024pitfalls}, as well as inability to deal with high-frequency robot data \cite{pertsch2025fast}. On the other hand, diffusion-based VLAs suffer from long training times \cite{zhang2024improving} \cite{wang2023patch}, and multiple denoising steps. Instead, \textit{EveryDayVLA} takes advantage of action chunking, parallel decoding, and a collaborative training recipe, predicting discrete autoregressive and continuous action chunks, using an L1-regression objective, to achieve high performance and inference rates. To further improve performance, we introduce a novel adaptive horizon module, allowing the module to execute less number of actions, when the two modes of actions stray from each others.

\subsection{EveryDayVLA}

\textit{EveryDayVLA} ingests as input a demonstration consisting of an image observation $o_{t}$ and a language instruction $l_{t}$. The model predicts an action $a_{t+1}$ to control the manipulator for the user-specified task.

\[
\pi : (o_t,\,l_t) \;\longrightarrow\; a_{t+1}
\]


The action \(a\) represents the end‐effector pose and is 7-DOF. This action vector includes:
\begin{itemize}
  \item 3-DOF for relative translation offsets: \([\Delta x,\,\Delta y,\,\Delta z] \in \mathbb{R}^3\),
  \item 3-DOF for rotation (Euler angles): \(\in \mathbb{R}^3\),
  \item 1-DOF for the gripper state (open/closed): \(\in \mathbb{R}^1\).
\end{itemize}

The ground truth (GT) and the predicted action lie in \(\mathrm{SE}(3)\), and are formulated as
\[
a = [\,\Delta x,\;\Delta y,\;\Delta z,\;\theta_x,\;\theta_y,\;\theta_z,\;0/1\,].
\]

\textbf{Vision-Language Model (VLM).} EveryDayVLA uses the Prismatic-7B VLM \cite{karamcheti2024prismatic}, which uses a two part visual encoder, containing pretrained SigLIP \cite{zhai2023sigmoid} and DinoV2 \cite{oquab2023dinov2} models. For the VLM, we leverage the Llama 2 language model backbone \cite{touvron2023llama}. To obtain continuous action outputs, we pass the final hidden states to a separate action head multi-layer perceptron (MLP). We get discrete action tokens (256-bin discretization of normalized actions) by again using the final hidden states, and performing softmax over the logits, to get a probability distribution over the action tokens.

\textbf{Collaborative Training.} We leverage a collaborative training recipe, inspired by \cite{liu2025hybridvla}, where we train a policy $\pi$ to predict continuous and discrete action chunks $a$ \(\in \mathbb{R}^{KXDX2}\), where $K$ and $D$ represent the size of the action chunk and the dimensionality of the end-effector pose respectively. To supervise the discrete actions, we use cross-entropy loss:

\begin{equation}
\mathcal{L}_{\mathrm{CE}}(\mathbf{A}_t,\hat{\mathbf{A}}_t)
  = -\sum_{k=1}^K A_{t,k}\,\log\bigl(\hat A_{t,k}\bigr).
\end{equation}

To supervise the continuous actions, we use the following L1-loss:

\begin{equation}
\mathcal{L}_{1}(\mathbf{A}_t,\hat{\mathbf{A}}_t)
  = \|\hat{\mathbf{A}}_t - \mathbf{A}_t\|_1\bigr.
\end{equation}

To simultaneously train continuous and
autoregressive action generation, we use the following combined loss function.

\begin{equation}
\mathcal{L}(\mathbf{A}_t,\hat{\mathbf{A}}_t)
  = \mathcal{L}_{\mathrm{CE}}(\mathbf{A}_t,\hat{\mathbf{A}}_t)
    \;+\;\lambda\,\mathcal{L}_{1}(\mathbf{A}_t,\hat{\mathbf{A}}_t)
\end{equation}

In practice, we set the weight $\lambda = 1$, to balance the optimization of discrete and continuous action outputs.


\textbf{Adaptive Horizon Ensembler.} Inspired by \cite{liu2025hybridvla}, we generate both discrete and continuous action chunks and observe that discrete actions excel at tasks requiring high‐level semantic reasoning, while continuous actions provide superior precision for fine manipulation. To harness both strengths, we introduce a more robust uncertainty metric: the mean absolute difference between the continuous and discrete action predictions, which outperforms the autoregressive model’s mean confidence score. Moreover, our collaborative training reveals that discrete actions outperform continuous outputs on average (Table \ref{tab:libero-results}). Accordingly, rather than fusing actions based on confidence thresholds \cite{liu2025hybridvla}, we adaptively adjust the planning horizon using our difference threshold—executing only those discrete action chunks whose mean absolute difference falls below this cutoff (see Algorithm \ref{alg:adapt_horizon_dplan}). We term our method \textit{AdaHorizon}, and find that it enables discrete actions to match or exceed the precision of continuous controls, even on dexterous tasks where discrete policies alone struggle. To satisfy real‐time constraints, we enforce a minimum execution length of four actions per eight‐step chunk.

\begin{algorithm}[ht]
\caption{Adaptive Horizon Ensembler}
\label{alg:adapt_horizon_dplan}
\begin{algorithmic}[1]
\Require Continuous actions $\{\mathbf{a}^c_t\}_{t=1}^T$, Discrete actions $\{\mathbf{a}^d_t\}_{t=1}^T$
\Require Parameters: $\mathrm{min\_actions}$, $\mathrm{replan\_threshold}$, $\mathrm{threshold}$, $\mathrm{max\_replan\_count}$, $\mathrm{next\_task\_thresh}$
\State \textbf{state}: $\mathrm{replan\_ctr},\,\mathrm{max\_replan\_ctr}$
\State $T \gets \text{length}(\mathbf{a}^c)$
\For{$t = 1 \to T$}
  \State $\mathrm{mad}_t \gets \frac{1}{D}\sum_{d=1}^{D}\bigl|\mathbf{a}^c_{t,d}-\mathbf{a}^d_{t,d}\bigr|$
\EndFor
\If{$\exists\,t\in[1,\dots,\mathrm{min\_actions}]\;:\;\mathrm{mad}_t > \mathrm{replan\_threshold}$ \textbf{and} $\mathrm{min\_actions}>1$}
  \State $\mathrm{replan\_ctr}\,\pluseq\,1$
\EndIf
\State $\mathrm{max\_replan\_ctr}\gets\max(\mathrm{max\_replan\_ctr},\;\mathrm{replan\_ctr})$
\If{$\mathrm{max\_replan\_ctr}\ge\mathrm{max\_replan\_count}$ \textbf{and} $\mathrm{replan\_ctr}\ge\mathrm{next\_task\_thresh}$}
  \State \Return $\{\mathbf{a}^d_t\}_{t=1}^T,\;\mathrm{mad}$
\EndIf
\State $\mathrm{mad\_mask}_t \gets (\mathrm{mad}_t < \mathrm{threshold})\quad\forall\,t$
\State $\mathrm{horizon}\gets \mathrm{min\_actions}$
\For{$t = \mathrm{min\_actions}+1 \to T$}
  \If{not $\mathrm{mad\_mask}_t$}
    \State \textbf{break}
  \EndIf
  \State $\mathrm{horizon}\gets t$
\EndFor
\State \Return $\{\mathbf{a}^d_t\}_{t=1}^{\mathrm{horizon}}$
\end{algorithmic}
\end{algorithm}

\subsection{Hardware}
\begin{figure}[t!]
    \centering
\includegraphics[
                     height=0.35\textheight,%
                     keepaspectratio]{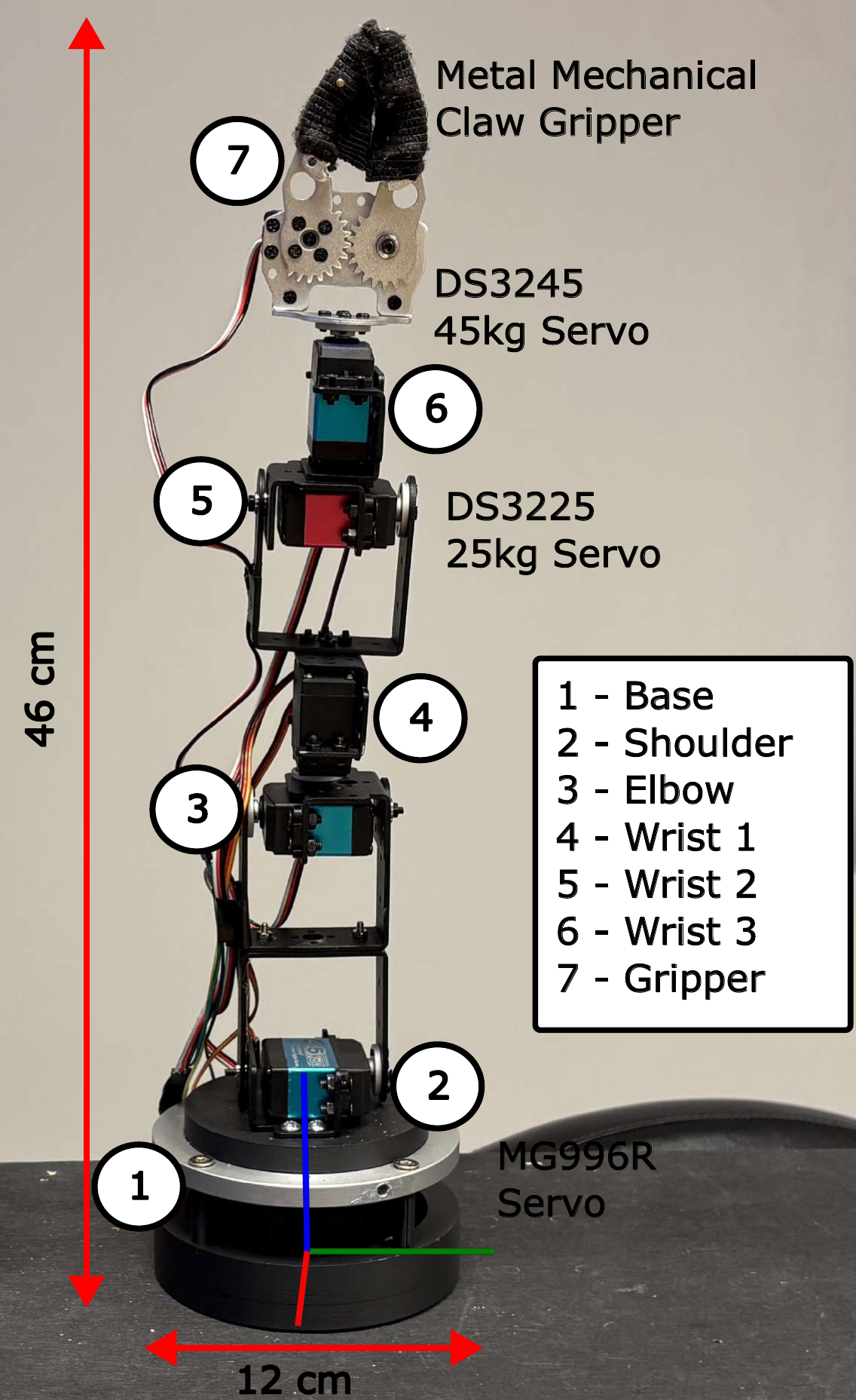}
\caption{
\textit{EveryDayVLA} hardware. The robot consists of 7 joints, including a base and claw gripper as the end-effector. In sum, the hardware costs \$311.98, affording 6 DOF, a payload of 0.2 kg, 382 mm reach, a max speed of 0.7 m/s and a repeatability of withing 10 mm.
}
\vspace{-.7cm}
\label{fig:hardware}
\end{figure}

We built a \$300 6-DOF robotic manipulator (see Figure \ref{fig:hardware}), using brackets on online marketplaces due to cost constraints. The robot arm is 38.2 cm from the base to the wrist and 46 cm from the base to the tip of the gripper. The robot is actuated by several types of off-the-shelf servo motors (MG996R, DS3225, DS3245) depending on the torque necessary for each joint. The 6 servomotors in the arm are configured in a roll-pitch-pitch-roll-pitch-roll configuration. The 3 servomotors comprising the wrist approximate a spherical joint. This configuration was chosen as the minimum number of servos to achieve any position or orientation. 

The bottom-most servo motor was held in a custom, 3-D printed base. The base was secured to 1/2 inch plywood board cut to 0.5m (length) x 1.0m (width) to be centered along the width and at the back of the board lengthwise, to maximize the manipulator's task space. 
Atop the base, we mounted a 120 mm swivel ball bearing. This ball bearing was secured to a custom 3-D printed plate where the bracket structure was then mounted.

To precisely control the movements of the robot, servo control, we use an off-the-shelf PCA9685 16-channel, 12-bit servo motor driver breakout board that communicates over I2C with an Arduino Uno. 




\section{Experiments}

\begin{table*}[h]
\centering
\caption{LIBERO task performance results. The best scores are highlighted in bold, and the second-best scores are underlined.}
\label{tab:libero-results}
\resizebox{\textwidth}{!}{
\begin{tabular}{lccccc}
\toprule
{\textbf{Policy inputs: third-person image, language instruction}} \\ 
\midrule
& \textbf{Spatial SR (\%)} & \textbf{Object SR (\%)} & \textbf{Goal SR (\%)} & \textbf{Long SR (\%)} & \textbf{Average SR (\%)} \\
\midrule
Diffusion Policy (scratch) \cite{chi2023diffusion} & 78.3 & 92.5 & 68.3 & 50.5 & 72.4 \\
Octo (fine-tuned) \cite{team2024octo} & 78.9 & 85.7 & 84.6 & 51.1 & 75.1 \\
DiT Policy (fine-tuned) \cite{hou2024diffusion} & 84.2 & 96.3 & 85.4 & 63.8 & 82.4 \\
OpenVLA \cite{kim2024openvla} & 84.7 & 88.4 & 79.2 & 53.7 & 76.5 \\
OpenVLA-OFT \cite{kim2025design} & 96.2 & \textbf{98.3} & \textbf{96.2} & \textbf{90.7} & \textbf{95.3} \\
Ours (Cont-L1 + AR, Use Cont-L1) & 95.0 & 92.4 & 90.0 & 79.4 & 89.2 \\
Ours (Cont-L1 + AR, Use AR) & \underline{96.4} & 95.0 & 90.0 & 80.8 & 90.6 \\
Ours (Cont-L1 + AR, Use \textit{AdaHorizon}) & \textbf{96.8} & \underline{95.6} & \underline{91.0} & \underline{82.0} & \underline{91.4} \\
\midrule
\end{tabular}
}
\end{table*}

\begin{figure*}
    \centering
\includegraphics[width=1.0\textwidth]{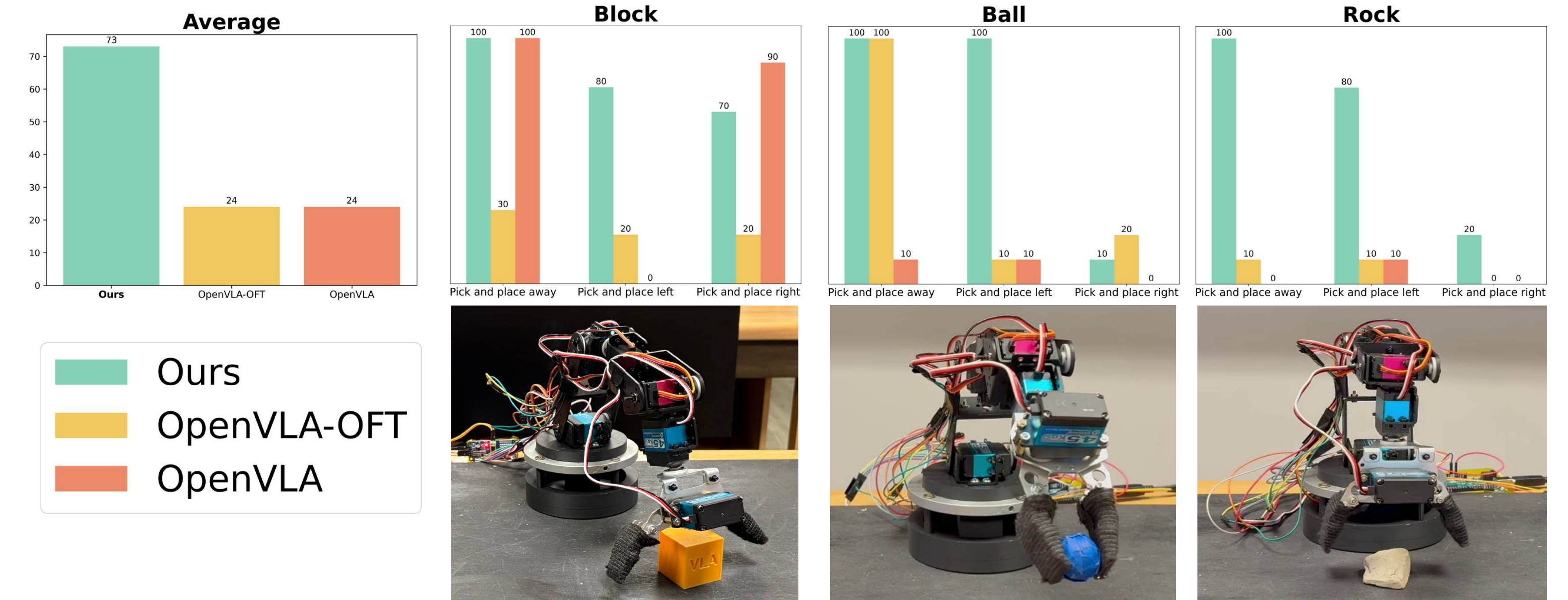}
\caption{Real-world evaluation results on in-distribution tasks, including picking a block, ball and rock. Our model is able to beat state-of-the art models on tasks and environments present in the training set by 49\% on average. We evaluate on three different objects, with three instructions each, "pick and place \{away, left, right\}". Our experiments show better success rates on almost every single task.}
\label{fig:real_world_eval}
\end{figure*}

\subsection{Dataset}


We fine-tune the OpenVLA-7B model \cite{kim2024openvla} on our custom dataset of 1,200 demonstrations—each pairing a natural-language instruction with an RGB observation sequence and corresponding end-effector poses—collected using our \$300 manipulator. Our dataset was captured across multiple tabletop environments and span a diverse task set: pick-and-place, environment manipulation (e.g., drawer opening/closing), and block-stacking. To streamline data collection, we abstracted core skills into parameterized trajectory primitives and paired them with generic language templates such as “pick up the ball and place it away from the robot,” enabling rapid generation of varied, yet semantically consistent, training samples.

\subsection{Implementation Details}


We extended the \cite{kim2025fine} codebase to jointly train autoregressive and regression heads and deploy with our AdaHorizon ensembler. For finetuning, we use LoRA \cite{lora} and 100 k iterations for simulation (on two A100s) and 50 k for real-world (on one A100). We use a LoRA rank of 32, batch size 8, and 4-step gradient accumulation.


We use IKPy \cite{Manceron_IKPy} for inverse kinematics to get robot joint angles. For the camera, we use an iPhone 12 mini and the DroidCam app \cite{DroidCam} to stream visual inputs to the model.

\subsection{Baselines}


On the LIBERO simulation benchmark, we report success rates across all four task suites, comparing against Diffusion Policy \cite{chi2023diffusion}, Octo \cite{team2024octo}, DiT Policy \cite{hou2024diffusion}, OpenVLA \cite{kim2024openvla}, and OpenVLA-OFT \cite{kim2025fine}. In real-world trials, we evaluate both in-distribution and out-of-distribution scenarios against OpenVLA and OpenVLA-OFT. We also measure inference throughput in simulation. Finally, we benchmark our adaptive-horizon ensemble, AdaHorizon, against ACT \cite{zhao2023learning}, HybridVLA \cite{liu2025hybridvla}, and COGAct \cite{li2024cogact}. 

For all experiments, we use the discrete actions, with the adaptive horizon ensembler (\textit{AdaHorizon}), which yields higher success rates and improved grasp accuracy compared to continuous-action baselines.

\section{Results}

\begin{table}[ht]
  \centering
  \caption{Inference rates on LIBERO.}
  \label{tab:inference-rate}
  \begin{tabular}{@{}l c@{\hspace{10pt}} c@{\hspace{2pt}}}
    \toprule
    \textbf{Method}       & \textbf{Inference Rate (Hz) $\uparrow$} & \textbf{Latency (sec) $\downarrow$}\\ 
    \midrule
    OpenVLA \cite{kim2024openvla}               & 4.2 & 0.2396                 \\ 
    OpenVLA-OFT \cite{kim2025fine}           & \textbf{109.7} & \textbf{0.0729}              \\
    Oursli         & \underline{54.2--108.4} & \underline{0.0738}  \\ 
    \bottomrule
  \end{tabular}
\end{table}




\begin{figure}
\centering
\includegraphics[width=0.4\textwidth]{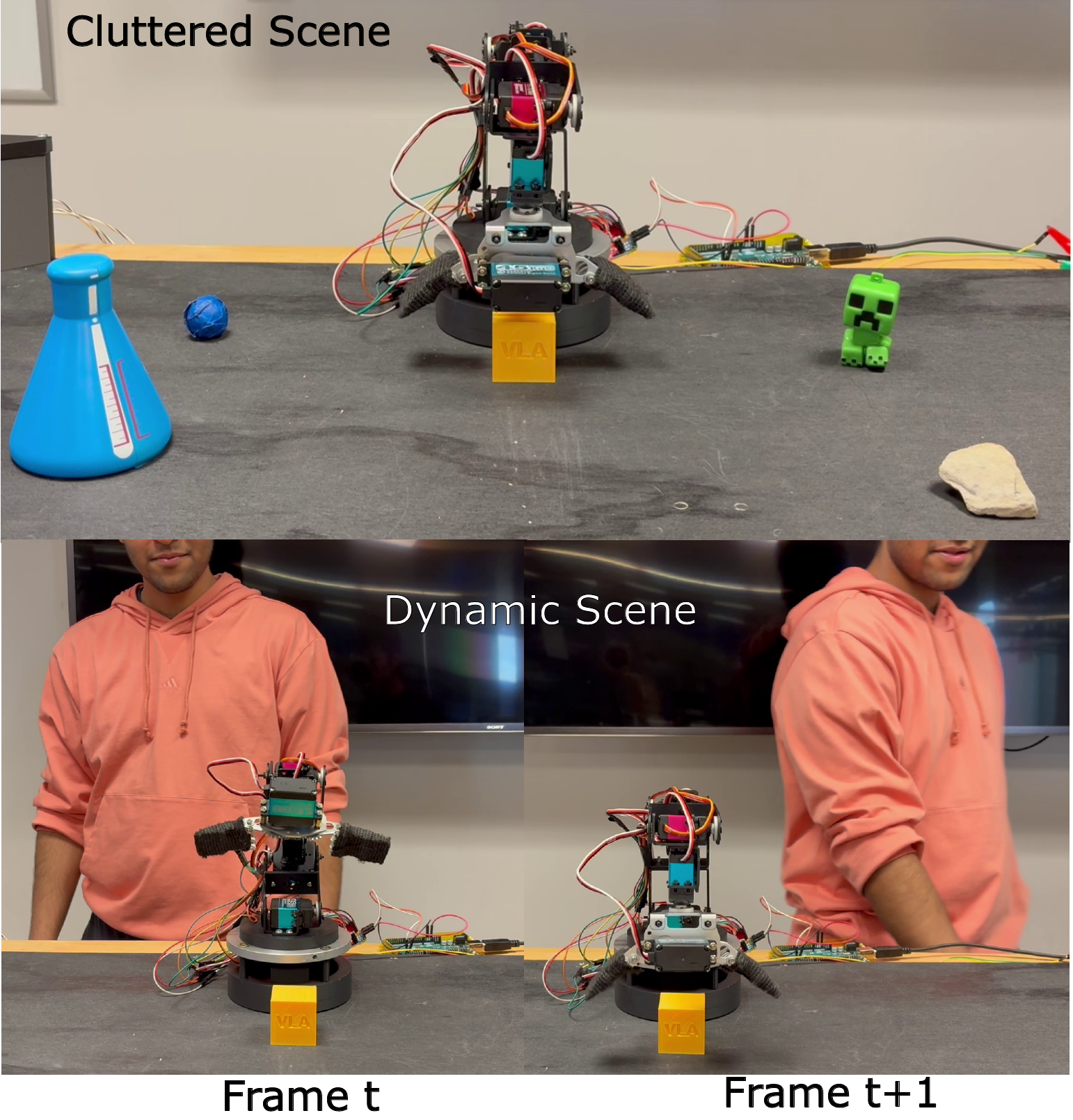}
\caption{Static and dynamic distractors. \underline{\textbf{Top:}} We benchmark our model with static distractors, and a cluttered scene where we add different objects, and vary the arrangement after every single trial. \underline{\textbf{Bottom:}} We benchmark with dynamic distrators, where a human walks in the scene and moves to distract the model.}
\label{fig:distractors}
\end{figure}

\begin{table}[h!]
  \centering
  \caption{Generalization and robustness to unseen tasks, environments and conditions}
  \label{tab:generalization}
  \begin{tabular}{l@{\hspace{1pt}} c@{\hspace{1pt}} c@{\hspace{1pt}} c@{\hspace{1pt}}}
    \toprule
      & Ours & OpenVLA-OFT \cite{kim2025fine} & OpenVLA \cite{kim2024openvla} \\
    \midrule
    Original   & \textbf{100}                & 30   &  \textbf{100}          \\
    OOD Tasks     & \textbf{90}\,(-10\%)       & 43\,(+43\%)   &  43\,(-57\%)   \\
    OOD Environments & \textbf{67.5} (-32.5\%)       & 20\,(-33\%) &     15\,(-85\%)   \\
    Static distractors     & \textbf{80}\,(-20\%)       & 30\,(+0\%)  &  70\,(-30\%)   \\
    Dynamic distractors  & \textbf{90}\,(-10\%)       & 20\,(-33\%)   &  60 (-40\%)   \\
    \bottomrule
  \end{tabular}
\end{table}

\begin{table}[h!]
  \centering
  \caption{Comparison of Action Ensemblers vs full discrete and continuous actions chunks as a baseline on LIBERO Spatial}
  \label{tab:ensemblers_comp}
  \begin{tabular}{l@{\hspace{3pt}} c@{\hspace{3pt}}}
    \toprule
      Method & Success rate (\%)\\
    \midrule
    ACT (Temporal Ensembling) \cite{zhao2023learning}  & 95.2  \\
    HybridVLA ($\theta = 0.8$) \cite{liu2025hybridvla}   & 94.2  \\
    COGAct (Use Cont-L1) \cite{li2024cogact}     & 93.6 \\
    Ours (Cont-L1 + AR, Use Cont-L1) & 95.0 \\
    Ours (Cont-L1 + AR, Use AR) & 96.4 \\
    Ours (Cont-L1 + AR, Use AdaHorizon)  & \textbf{96.8} \\
    \bottomrule
  \end{tabular}
\end{table}

\subsection{Results on LIBERO simulation benchmark}

Across the four LIBERO suites, our method trails the top-performing baseline by an average of 3.9 \% (Table \ref{tab:libero-results}). Notably, we outperform the best method on the Spatial suite by 0.6 \%, while incurring drop-offs of 2.7 \%, 5.2 \%, and 8.7 \% on the Object, Goal, and Long suites, respectively.

In an ablation against using only continuous (Cont-L1) or only discrete (AR) actions, \textit{AdaHorizon} consistently improves performance in every suite—on average by 0.8 \% over the better of the two single-modality policies, demonstrating the value of adaptively combining both action outputs.

We also achieve an inference rate of up to 108.4 Hz (Table \ref{tab:inference-rate}), adding only 0.9 ms of latency relative to OpenVLA-OFT \cite{kim2025fine}, which corresponds to the overhead of our adaptive-horizon module.

\subsection{Results on Real-World Tests}

In real-world, in-distribution pick-and-place experiments, EverydayVLA outperforms other methods by an average of 49\% in success rate across blocks, balls, and rocks (Figure \ref{fig:real_world_eval}). Although our model performs similarly to \cite{kim2024openvla} on picking and placing blocks (most common task in the collected dataset), the latter struggles when placing to the left, and often fails to release the block (once grasped) in a timely manner. On ball pick-and-place, \textit{EverydayVLA} exceeds both baselines in every variant except “pick-and-place right.” For rock manipulation, our model achieves comparable success, underscoring its robustness across diverse object types.

The primary failure mode for \textit{EverydayVLA} is delayed object release and not finishing the task in a timely manner. OpenVLA-OFT \cite{kim2025fine} most often fails due to misaligned grasps and excessive current draw (which forces a safety shutdown), while OpenVLA \cite{kim2024openvla} suffers from intermittent stuttering and pauses that prevent task completion within the allotted time.

\subsection{Generalization results}

On generalization and robustness evaluation of unseen tasks, environments and conditions (Table \ref{tab:generalization}), our model does the best. We outperform the OpenVLA-OFT \cite{kim2024openvla} and OpenVLA \cite{kim2025fine}, on generalization to unseen tasks, environments, and robustness to static and dynamic distractors. We notice minimal dropoffs in presence of distractors. 

Compared to static distractors, such as a more cluttered environment, our model only experiences a 20\% decline in performance compared to the training environment. Further, in presence of dynamic, moving distractors, such as humans, we see only a 10\% performance dropoff.

\subsection{Results on Comparison to Action Ensemblers}

We evaluate our ensembler against competing methods on the LIBERO Spatial suite (Table \ref{tab:ensemblers_comp}). \textit{AdaHorizon} outperforms all baselines by dynamically adjusting the executed action‐chunk length: for complex manipulation tasks, shorter chunks enable timely replanning and higher success rates, whereas for simpler tasks full‐chunk execution is optimal.
Almost all existing ensemblers, except temporal ensembling \cite{zhao2023learning}, fall below even the continuous and discrete single‐modality baselines. In particular, temporal ensembling, HybridVLA \cite{liu2025hybridvla}, and COGAct \cite{li2024cogact} oversmooth the action stream via sliding‐window aggregation, sacrificing the fine‐grained control needed for dexterous tasks. Our ensembler shows a 1.6\% improvement in success rate compared to the next best method.


\section{Conclusions}

In this work we present \textit{EveryDayVLA}, a framework with a low-cost manipulator, a VLA leveraging a novel collaborative training and an adaptive horizon (\textit{AdaHorizon}) that computes uncertainty from the predicted discrete and continuous actions, adaptively modifying the action horizon. 

Our work demonstrates substantial improvement on in-distribution and out-of-distribution scenarios. Specifically, on in-distribution scenarios our model showcases 49\% improvement on average compared to the second-best method. On out-of-distribution scenarios, including unseen tasks, environments, static and dynamic distractors, \textit{EveryDayVLA} beats the next best method by 34.9\% on average. Further, we show competitive results to the LIBERO benchmark, placing second-best on average success rate. We also show that our model is fast and that its highest inference rate nearly matches that of \cite{kim2025fine}. Finally, we beat other existing action ensemblers with our Adaptive Horizon action ensembler, beating the second-best method by 1.6\% on the LIBERO Spatial task suite.

Our work has a few limitations. On the hardware front, we have not ensured long-term robustness. 
In the future, we look to improve the arm's mechanical durability. We also experience limitations in executing fine-grained manipulation, which is due to the limited servo precision as well as relatively low number of expert demonstrations in our fine-tuning dataset (compared to simulation). An extension of our work is to use higher precision servos, and collect more expert trajectories to consolidate our dataset for improved fine-grained control.







\section*{ACKNOWLEDGMENT}

We would also like to thank the Senior Design course professors and teaching assistants for their input on developing the PCB. We would like to thank the staff at Student Electronics Resource Center for assistance with providing the Arduino micro-controller.

\bibliographystyle{IEEEtran}
\bibliography{literature}

\end{document}